\documentclass[conference]{IEEEtran}
\IEEEoverridecommandlockouts
\usepackage{cite}
\usepackage{amsmath,amssymb,amsfonts}
\usepackage{algorithmic}
\usepackage{graphicx}
\usepackage{textcomp}
\usepackage{xcolor}
\usepackage{graphicx}
\usepackage{subfig}
\usepackage{lipsum}
\usepackage{authblk}

\IEEEoverridecommandlockouts
\newcommand{\subparagraph}{This work is supported by grant}

\usepackage{titlesec}
\def\BibTeX{{\rm B\kern-.05em{\sc i\kern-.025em b}\kern-.08em
    T\kern-.1667em\lower.7ex\hbox{E}\kern-.125emX}}

\begin{document}
\titlespacing{\subsection}{0pt}{\parskip}{-\parskip}
\titlespacing\section{0pt}{2pt plus 4pt minus 2pt}{0pt plus 2pt minus 2pt}
\titlespacing\subsection{0pt}{2pt plus 4pt minus 2pt}{1pt plus 2pt minus 2pt}
\title{Heterogeneous Learning from Demonstration\\
\thanks{Research supported by MIT Lincoln Laboratory under grant 7000437192.}
% \thanks{Supported by grant no. 7000437192 from MIT Lincoln Laboratory.}
}

\author{Rohan Paleja and Matthew Gombolay\\
\text{Georgia Institute of Technology} \\
\text{rpaleja3@gatech.edu, matthew.gombolay@cc.gatech.edu}}

% \author{\IEEEauthorblockN{1\textsuperscript{st} Rohan Paleja}
% \IEEEauthorblockA{\textit{Institute of Robotics and Intelligent Machines} \\
% \textit{Georgia Tech}\\
% Atlanta, Georgia \\
% rpaleja3@gatech.edu}
% \and
% \IEEEauthorblockN{2\textsuperscript{nd} Matthew Gombolay}
% \IEEEauthorblockA{\textit{Institute of Robotics and Intelligent Machines} \\
% \textit{Georgia Tech}\\
% Atlanta, Georgia \\
% matthew.gombolay@cc.gatech.edu}
% }

\maketitle

\begin{abstract}
The development of human-robot systems able to leverage the strengths of both humans and their robotic counterparts has been greatly sought after because of the foreseen, broad-ranging impact  across industry and research. We believe the true potential of these systems cannot be reached unless the robot is able to act with a high level of autonomy, reducing the burden of manual tasking or teleoperation. To achieve this level of autonomy, robots must be able to work fluidly with its human partners, inferring their needs without explicit commands. This inference requires the robot to be able to detect and classify the heterogeneity of its partners. We propose a framework for learning from heterogeneous demonstration based upon Bayesian inference and evaluate a suite of approaches on a real-world dataset of gameplay from StarCraft II. This evaluation provides evidence that our Bayesian approach can outperform conventional methods by up to 12.8$\%$.

\end{abstract}

\begin{IEEEkeywords}
Learning from Demonstration; Human-Robot Interaction; Human-Robot Teaming; Deep Learning
\end{IEEEkeywords}

\section{Introduction}
\par To achieve a fluent human-robot teaming, there must be a precise balance between manual control by the human and autonomy for the robot. Both extremes -- the need to manually task all robot activity versus fully autonomous robots -- have been shown to degrade overall performance~\cite{Chen:2012,844354}. Finding this exact balance is difficult as a robot working in a collaborative setting requires the ability to anticipate and adapt to its human partner. Because of the significant degree of difference (i.e., heterogeneity) between individual humans, learning to tailor the robot behavior to each human is intractable with conventional techniques.

\par The goal of this work is for a robot to contribute as a high performing teammate, learning to anticipate the needs/actions of its human partners. To accomplish this goal, the robot must develop and maintain a joint mental model of each demonstrator's policy while taking into account that there may be significant differences amongst the robot's various human teammates. For example, consider a robot acting as a scrub nurse that has the job of handing a surgeon a tool during a procedure; given a large dataset of surgeons' preferences for a tool during a certain procedure, the robot can attempt to hand the surgeon the correct tool rather than burdening the surgeon with having to explicitly task the robot to fetch the desired instrument.

\begin{table*}[t]
  \centering
\begin{tabular}{|p{.8cm}|p{.8cm}|p{3.5cm}|p{6.2cm}|p{.8cm}|p{2cm}|}
\hline
Method & $f_{NN}$ & $f_{NN(i)}$  & $f_{BNN}$ &  $f_{LSTM}$    & $f_{B-LSTM}$                       \\ \hline
Train  & All data  & On each $\textit{i}^{th}$  data cluster, created through k-means clustering \cite{Nikolaidis:2015:EML:2696454.2696455} 
& All data; Start of Training: Trains as $f_{NN}$; End of Training: $\omega$ and a fully connected layer are appended. Only the weights of the F.C. layer and $\omega$ are tuned. & All data      & All data     \\ \hline
Test   & Weights fixed & Weights fixed  &                                                                                        Same test scheme used in $f_{BNN}$                                                                  & Weights fixed & Same test scheme used in $f_{BNN}$ \\ \hline
\end{tabular}
\title{}
  \caption{Neural Networks and their Train/Test Schemes}
  \label{tab:1}
  \vspace{-6mm}
\end{table*}

\par While this task seems straightforward, the disparities between surgeons' preferences makes it difficult to learn from the data. Early work in Learning from Demonstration (LfD) found that pilots executing the same flight plan created such variance in the data as to make it more practical to learn from a single pilot and disregard the remaining data \cite{DBLP:conf/icml/MoralesS04}. Sammut et. al. \cite{DBLP:conf/icml/SammutHKM92} showed that when attempting LfD from pilots' demonstrations executing a single flight plan, averaging trajectories led to worse performance than using a single trajectory. Nikolaidis et al. \cite{Nikolaidis:2015:EML:2696454.2696455} approached this issue by categorizing demonstrators according to their task execution preference by clustering and learning a separate policy for each cluster. While this allows for utilization of the entire dataset, each policy learns off a fraction of the data. Returning to the robotic scrub nurse scenario, this clustering method would split the data into categories based on the type of surgeon, such as intern, resident, or attending surgeon, and learn a separate policy for each type based upon only the data representing the respective type. We hypothesize that a robot scrub nurse would have a better estimate of a surgeon's preferred surgical workflow if the robot were to reason about the data within that surgeon's cluster (e.g., all resident surgeons' data given the robot's partner is a resident) as opposed to treating all surgeons the same.  While learning from clusters may provide preferences closer to optimality, it makes the learning problem harder by providing 1/$k^{th}$ of the data, where k is the number of clusters. Further, this cluster-based approach does not account for variability among surgeons within the same cluster. 

\par Instead, we believe that accounting for the commonalities and differences amongst demonstrators within a cluster (or across all clusters) would allow for a more complete use of the data. Our method seeks to estimate the demonstrator's ``style" or ``unique descriptor" (i.e., latent embedding) in real-time by employing a Bayesian Neural Network (BNN), which reasons about the discrepancy between the average demonstrator and the specific one currently being observed. In our case, this descriptor is represented by a low-dimensional latent encoding vector $\omega$, where the length of the vector is a hyper-parameter that can be indicative of the complexity of the style. $\omega$ explicitly synthesizes features to explain the variance of a demonstrator that isn't accounted for in the one-size-fits-all part of the network. In our network  $f_{BNN}$, $\omega$  can be learned in real-time via an auto-encoder \cite{HinSal06} or via backpropagation \cite{DBLP:conf/aaai/KillianKD17}, as we do in our proposed method.

We believe that exploration into heterogeneous learning from demonstration will allow for a large increase in human-robot task utility. In turn, this ability to learn a human teammate's behavior can be leveraged to give specific benefits in surgery, manufacturing, and search \& rescue. 

\section{Research Approach}
\par As a testbed, we utilize the StarCraft II API PySC2 \cite{vinyals2017starcraft}, which allows for analyzing real-player game replays. We use this testbed rather than proceed to human LfD since a large dataset of human demonstrations (i.e., 1-vs.-1 game replays) are readily available. StarCraft II poses an difficult challenge as it is a real-time, continuous-state-space, partially observable, strategy game. 

\subsection{Effects of Utilizing Heterogeneity}
\par We consider the problem of learning to mimic the decision-making of players (i.e., direct policy learning) within StarCraft II. We consider the following algorithmic formulations for comparison, as shown in Table 1. First, we consider a regular neural network $f_{NN}$, which is our baseline. Second, we include the clustering-based method, denoted $f_{NN(i)}$, by Nikolaidis et al. \cite{Nikolaidis:2015:EML:2696454.2696455} in which we cluster the gameplay data into three partitions using k-means and learn a separate policy network on each. Third, we consider a BNN, $f_{BNN}$, which is able to holistically reason about the homo- and heterogeneity amongst the demonstrators. 

\par Finally, we believe that there may be important, latent, time-varying information (e.g., phases of gameplay dynamics) that may need to be explicitly captured. The standard model for capturing these dynamics is a Long-/Short-Term Memory (LSTM) neural network. We augment these networks here to be able to reason about static heterogeneity amonsgt players by including a Bayesian encoding structure in a network we designate $f_{B-LSTM}$. We also include a generic LSTM ($f_{LSTM}$) as a baseline.

\section{Results and Discussion}\label{AA}

\par Table \ref{tal} provides promising evidence that our Bayesian-LSTM formulation, which captures static heterogeneity and time-varying gameplay phenomena, improves the performance of LfD mechanisms. In future work, we plan to conduct a sensitivity analysis to isolate robust hyperparameters, find better loss functions, and discover regularizers to isolate player-dependent features.

\begin{table}[h!]
\centering
\begin{tabular}{|c|c|c|c|}
\hline
$f_{NN(i)}$ & $f_{BNN}$ & $f_{LSTM}$ & $f_{B-LSTM}$ \\ \hline
 161.6 $\pm 2.1$\%   & 101.6 $\pm 1.5$\%   & 98.9  $\pm 1.7$\%   & \bf{87.2}  $\pm 1.5$\%      \\ \hline
\end{tabular}
\title{}
  \caption{Average Loss Normalized to $f_{NN}$ Loss}
  \label{tal}
\end{table}

\par
\begin{figure}[h!]
	\centering
	\includegraphics[width = \columnwidth]{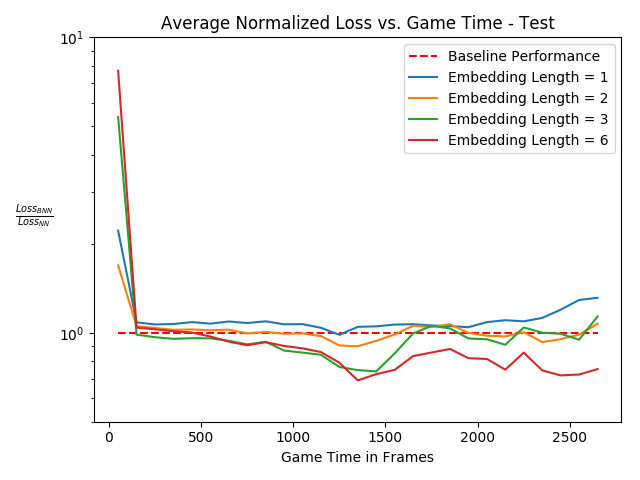}
	\caption{Performance Analysis of Encoding Lengths.}
	\label{fig:method}
\end{figure}

 Figure \ref{fig:method} depicts how the performance of the BNN changes with various encoding lengths and the effects of training on data that is clustered. It can be seen that as the game proceeds the $f_{BNN}$'s with encoding lengths of three and six clearly outperform the $f_{NN}$. This result supports the hypothesis that holistically reasoning about heterogeneity is helpful.

\section{Future Work}
\par Successful human-robot teaming requires robot algorithms that are able to take into account the heterogeneity of humans and allow the robot to tailor to the needs of its unique team. The destination of this work is to achieve a feat similar to that of a robot scrub nurse. Given demonstrations of a certain task performed, the robot can infer the demonstrator's style and assist him/her in the best way. The work presented in this paper provides insight into the development of a continuous mapping function between the encoding and the demonstrator policy. Moving forward, a complimentary robot policy must be identified that will result in a human-robot teaming setting of the highest performance.

\par Theoretical expansions upon this work include adding in an estimate of uncertainty of the current demonstrator style to inform active learning mechanisms and seeking alternate loss functions for tuning $\omega$ to maximize the information captured and resisting the tendency to overfit. 

\par We also note that recent techniques in meta-learning, e.g. Finn et. al. \cite{finn2017model}, have sought to learn a network that can be quickly tailored to perform well for a single task drawn from a known distribution of tasks. Once tailored to a specific task, this policy loses its ability to then be tailored to different tasks. In contrast, our approach is able to switch between tasks (e.g., predicting humans' actions) by adapting a relatively small vector encoding, $\omega$, rather than tuning the entire network. 

\par Further, human-subject experimentation can be performed to analyze the performance of heterogeneous LfD. A sample starting experiment would be to give a robot demonstrations of left-handed people and right-handed people performing some task, and ask it to identify the dominant hand of a current demonstrator. We can then move to test the performance of a robot learning a complimentary policy to assist those based on their distinctive style. Overall, these results would produce interesting inquiries into heterogeneous LfD and allow us to push the utility of human-robot teaming.

\newpage
\bibliographystyle{ieeetr}
\bibliography{reference}{}
\end{document}